\documentclass[10pt, journal, letterpaper]{IEEEtran}

\usepackage{cite}
\usepackage{amssymb}

\ifCLASSINFOpdf
   \usepackage[pdftex]{graphicx}
\else
   \usepackage[dvips]{graphicx}
\fi
\usepackage{subfigure}
\usepackage{multirow}
\usepackage[cmex10]{amsmath}

\usepackage{array}

\usepackage{url}
\usepackage{balance}

\usepackage{algorithmic}
\usepackage{algorithm}

\begin{document}
%
\title{Power Data Classification: A Hybrid of a Novel Local Time Warping and LSTM}
\author{Yuanlong Li,~\IEEEmembership{Member,~IEEE,}
        Han Hu,~\IEEEmembership{Member,~IEEE,}
        Yonggang Wen,~\IEEEmembership{Senior Member,~IEEE,}
        and Jun Zhang~\IEEEmembership{Senior Member,~IEEE}

\thanks{Yuanlong Li, Han Hu and Yonggang Wen are with School of Computer Engineering, Nanyang Technological University, Nanyang Avenue, Singapore 639798. Email: \{liyuanl, hhu, ygwen\}@ntu.edu.sg.}
\thanks{Jun Zhang is with School of Computer Science and Engineering, South China University of Technology, Guangzhou, China. Email: junzhang@ieee.org. }
}

\maketitle

\begin{abstract}
In this paper, for the purpose of data centre energy consumption monitoring and analysis, we propose to detect the running programs in a server by classifying the observed power consumption series. Time series classification problem has been extensively studied with various distance measurements developed; also recently the deep learning based sequence models have been proved to be promising. In this paper, we propose a novel distance measurement and build a time series classification algorithm hybridizing nearest neighbour and long short term memory (LSTM) neural network. More specifically, first we propose a new distance measurement termed as Local Time Warping (LTW), which utilizes a user-specified set for local warping, and is designed to be non-commutative and non-dynamic programming. Second we hybridize the 1NN-LTW and LSTM together. In particular, we combine the prediction probability vector of 1NN-LTW and LSTM to determine the label of the test cases. Finally, using the power consumption data from a real data center, we show that the proposed LTW can improve the classification accuracy of DTW from about 84\% to 90\%. Our experimental results prove that the proposed LTW is competitive on our data set compared with existed DTW variants and its non-commutative feature is indeed beneficial. We also test a linear version of LTW and it can significantly outperform existed linear runtime lower bound methods like LB\_Keogh. Furthermore, with the hybrid algorithm, for the power series classification task we achieve an accuracy up to about 93\%. Our research can inspire more studies on time series distance measurement and the hybrid of the deep learning models with other traditional models.

\end{abstract}



\IEEEpeerreviewmaketitle

\section{Introduction}
\IEEEPARstart{N}{owadays}, a growing number of data centres have been built to support complicated computation and massive storage required by various blooming applications \cite{DataCentreEnergyGrow}. Each data center is typically equipped with hundreds of thousands servers and requires many mega-watts electricity to power its hosted servers and the auxiliary facilities \cite{barroso2007case}. An essential problem is to monitor such a large amount of servers for energy saving and maintaining the business continuity.

Monitoring technologies \cite{moore2005data} can be divided into two categories: intrusive and non-intrusive. Intrusive technologies require the install of certain monitoring software which requires the administration role of the system. Compared to the intrusive methods, non-intrusive methods are more flexible, which only require limited data for the monitoring and analysis.

In this paper, for the purpose of energy consumption monitoring, we propose to detect the running program in a server by analysing the observed power consumption series. The power data can be measured without the administration right of the server, which can be useful in collecting the power related information of the servers for the purpose of energy consumption analysis. The proposed classification analysis can only gain the type of the running program, avoiding any possibility in accessing the privacy-related contents in the server.

The proposed program detecting problem falls into the field of time series classification. As a time series classification problem, the power data classification problem can be challenging as the power series collected in detection may be only a small piece of the whole power series of a program, with incomplete and limited information. For this problem, the key is to design an accurate and fast classification algorithm.

Currently there are a few similar works on classifying signals (like the power consumption signals studied here) such as \cite{fehske2005new} \cite{soliman1992signal} \cite{reaz2006techniques}. However, the technologies applied in these literature are based on common spectral or statistical features with classifiers such as nearest neighbour or neural network. In a more general aspect, the time series classification problem has been extensively studied \cite{bagnall2016great}, among which the most popular method is 1-nearest neighbour with  dynamic time warping (DTW). The major research line in time series classification has been the developing of various DTW based distance measurements (variants such as \cite{stefan2013move} and enhancers \cite{batista2011complexity}); yet we find that even though these measurements can be better than the original DTW by certain degree for certain cases, these variants all have been designed to incorporate the dynamic programming idea of DTW (except some lower bound methods like LB\_Keogh \cite{rakthanmanon2012searching}) and all designed to be commutative.  Another line of research has also become notable recently, i.e. the long short term memory (LSTM) neural network, which shows great modelling ability for sequential data. In this work, we propose a novel classifier with much higher accuracy and based on the great efforts in the current literature.

In this research, firstly, we propose a Local Time Warping (LTW) time series distance measurement, which is a light weight DTW variant that does not need the dynamic programming procedure and is designed to be non-commutative. LTW can be set to a linear runtime algorithm which can perform almost as good as the DTW on our data set. Secondly, instead of further enhancing the distance measurement which can be much more complicated and time consuming, we look into a less expensive solution, which is to develop a hybrid algorithm of the 1-nearest neighbour with LTW (1NN-LTW) and the recent deep learning model for time sequential modelling. To do so, we first utilize the state-of-art sequential data modelling neural network LSTM \cite{hochreiter1997long} \cite{gers2000learning} to classify the power series. Then we propose a new hybrid algorithm of the proposed 1NN-LTW and the LSTM. Our study shows that both 1NN-LTW and LSTM can outperform the 1NN-DTW with similar accuracy; however, these two algorithms have their unique different natures and the accurately classified samples of these two algorithms have significant differences. The hybrid algorithm of the two classifiers, termed as LSTM/LTW, improves the classification accuracy further easily.

{The main contributions of this paper are summarized as follows: }
\begin{itemize}

   \item We propose a new distance measurement LTW. LTW has two unique features which are different from the existing DTW variants: 1) LTW is based on simple ``local warping'', no dynamic programming procedure is needed; 2) LTW is non-commutative and is flexible for the nearest neighbour classifier for the time series classification problem. Our experimental results show that for our problem, the proposed LTW can perform better than DTW and its several different variants. Also our experiment shows that the non-commutative feature of LTW is beneficial. Furthermore, the linear version of LTW can perform almost as good as DTW on our data set. These results show that for certain cases, a light weight local warping distance measurement (such as the LTW) may be good enough for the classification task; however, this does not mean that the proposed LTW can work for all kinds of time series data sets.

  \item For the first time, we develop a hybrid algorithm of 1NN-DTW and LSTM termed as LSTM/LTW. The hybrid algorithm is based on a well trained LSTM neural network. Although the training procedure of the LSTM can be time consuming, the classification process can be fast in testing with the LTW distance.

  \item Numerical experiments show that for the power data classification problem, with the LTW distance measurement, the accuracy of the 1NN-LTW classifier can be improved from about 84\% to about 90\% compared to the 1NN-DTW. With the hybrid algorithm LSTM/LTW, we achieve the power consumption series classification accuracy upto about 93\%, which proves that using the power consumption series to detect the type of the running programs in a server can be very accurate.
\end{itemize}

The remainder of this paper is organized as follows. In Section \ref{stn:related}, we briefly introduce the state-of-art time series classification algorithms. In Section \ref{stn:data_collection_ana}, we introduce the experimental data collection design and some preliminary analysis on the data. In Section \ref{stn:new_algorithm}, we introduce the new proposed algorithm and in Section \ref{stn:Numer} we show the numerical evaluation results and the analysis. In Section \ref{stn:summary} we conclude the whole paper and introduce the future works.


\section{Related Works}
\label{stn:related}
The power data classification problem studied in this paper can be taken as a time series classification problem, which has been studied extensively for the past decades. For this problem, common classifiers like support vector machine (SVM), $k$-nearest neighbour (KNN) with Euclidean distance have been proved to be non-competitive to the DTW distance measurement based method like 1NN-DTW \cite{xi2006fast}. Recently there have been a lot of new methods which have been proved to be as competitive as 1NN-DTW. On one hand, there are many non-neural network based methods like Shapelet based method, dictionary based methods, interval based methods and ensembles of these methods. We will brief these methods below. On the other hand, recently with the fast development of deep learning \cite{hinton2006reducing}, LSTM neural network has also been proved to hold high modelling ability for sequential data. In the following we will briefly introduce LSTM.

\subsection{Non-Neural Network Approaches}
The most popular non-neural network time series classifiers are nearest neighbour based method with various different distance measurements. The most popular method is the 1NN-DTW, which is a special $k$-nearest neighbour classifier with $k=1$ and a special DTW distance measurement. For the 1-NN classifier, the common standard procedure to label of a test sample given a set of training samples is as follows. First the distances of the test sample to all the training samples are computed; then the training sample that has the smallest distance to the test sample is chosen and its label is assigned to the test sample as the classification result. In the above procedure, the key is to utilize a proper distance measurement. For 1NN-DTW, the DTW distance is used, which has superior performance for time series data.

\begin{figure}[!t]
\centering
   \includegraphics[width=0.7\columnwidth] {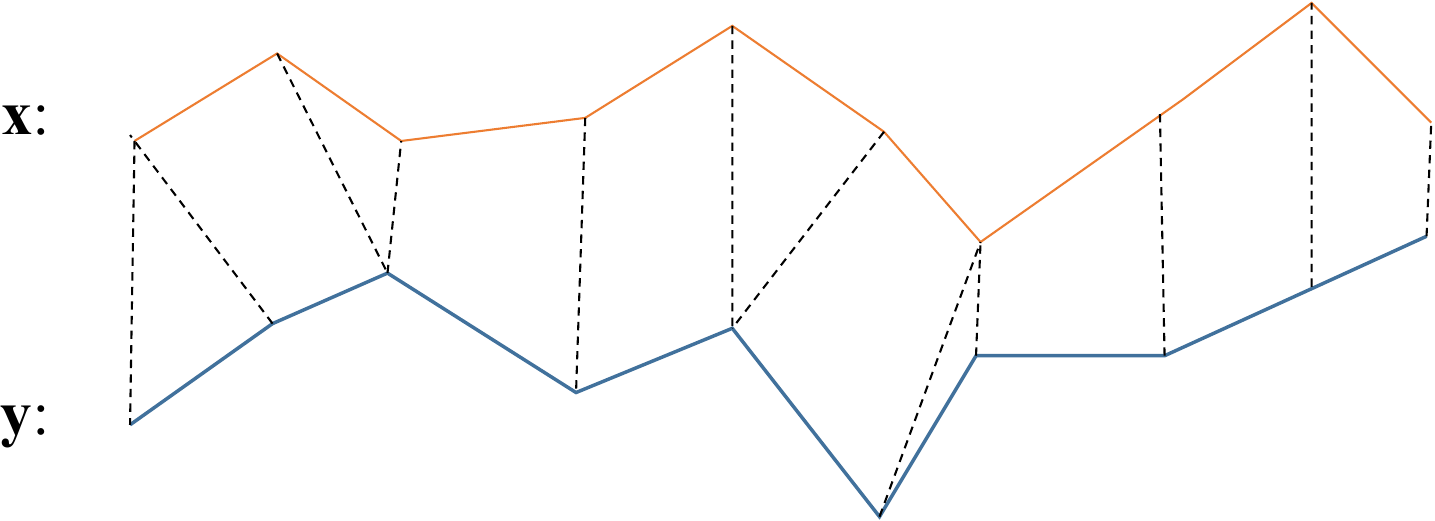}

\caption{Illustration of the DTW distance measurement. In computing the DTW distance between samples $\mathbf{x}$ and $\mathbf{y}$, the DTW algorithm finds the best match (shown by the dash lines) between the two series at different time steps.}
\label{fig:DTW}
\end{figure}

The DTW calculates the distance of two sequences $\mathbf{x}$ and $\mathbf{y}$ in a manner of finding the best match between them, as shown in Fig. \ref{fig:DTW}. The idea is that sequential data often contain similar fluctuation patterns, however, a same pattern, when existed in different sequences as sub-sequences, may be stretched, shrank or delayed in the time axis. In this case, the DTW distance measurement aims to warp the time axis non-linearly and finds the best match between the two samples such that when a same pattern exists in both sequences, the distance is smaller.

Mathematically, the DTW distance is computed by the following dynamic programming process. Denote $D(i,j)$ as the DTW distance between sub-sequences $x[1:j]$ and $y[1:j]$, then the DTW distance between $\mathbf{x}$ and $\mathbf{y}$ can be computed by the dynamic programming process with the following iterative equation:
\begin{equation}
D(i,j)=min\{D(i-1,j-1), D(i-1,j), D(i,j-1)\}+|x_i-y_j|.
\end{equation}
The time complexity to compute the DTW distance is $O(nm)$, where $n$ and $m$ are the length of $\mathbf{x}$ and $\mathbf{y}$ respectively. The DTW distance measurement actually re-align the time step index pairs in the computing of the distance. In practice, usually a threshold $w$ is used to restrict the index offset in the alignment, which can be critical to the classification results \cite{ratanamahatana2004everything}. Also there are many study \cite{salvador2007toward} working on accelerating the computing speed of DTW, which results in the fast DTW that can be computed in linear time of the length of the sequences. In this paper, we follows the idea of DTW but propose a new distance measurement, which can be computed with a local warping index set without a dynamic programming process and has a special non-communicative nature that can be helpful.

There are many DTW variants proposed. We name only a few here for the space constraint; one can refer to \cite{bagnall2016great} for a more complete review and comparison of the existing methods. Move-Split-Merge \cite{stefan2013move} introduces move and split operation in dynamic warping. Complexity Invariant distance (CID) \cite{batista2011complexity} is a weight modifier which can be used to enhance any kind of distance measurement, and is proved to be very useful when using with DTW.

Besides DTW based method, there have been many new different methods which look into the pattern of the time series for classification. For example, Shapelet \cite{rakthanmanon2013fast} based methods utilize the subsequences that can differentiate different classes to do the classification. Dictionary based methods \cite{lin2007experiencing} transforms the series into discrete words in a dictionary and then do the classification. Interval Based Classifiers \cite{deng2013time} try to extract the feature from intervals in each time series for the classification. In this paper, we will focus on DTW based methods and will not compare with these methods, as proved in \cite{bagnall2016great}, these methods perform similar with DTW based methods unless they are ensemble methods. Ensemble methods are the kind of classifier that combine multiple simple classifiers which can be better than any single classifier. Currently, the existed ensemble methods are mainly based the above listed classifiers, and we have not seen any work on ensemble method of the above methods and neural network. In this paper, we will thus propose a simple hybrid algorithm of a nearest neighbour classifier and neural network. The neural network classifier used in this paper is introduced below.

\subsection{LSTM}
LSTM is first proposed by Hochreiter and Gers et al. as an upgrade of the recurrent neural network (RNN) \cite{mikolov2010recurrent}. RNN is used to handle sequential data with a special calculation process following the time step increment, while traditional neural network simply treats the sequence as a plain vector. With such nature, RNN is suitable for modelling sequential data. However, it suffers from a problem called diminishing gradient, which is caused by the iterative process on the time axis and makes the gradient used in the training process extremely small and causes training failure. To solve the problem, the LSTM is proposed and it utilizes a memory core to avoid the diminishing gradient. The details of the LSTM neural network will be introduced in Section \ref{stn:new_algorithm}.

LSTM has shown great modelling power for sequential data and has been successfully applied in various machine learning fields like natural language process (NLP) \cite{donahue2015long}, video analysis \cite{yue2015beyond} and etc. It is also noted that LSTM can be both discriminative and generative. By discriminative, LSTM can be used for classification tasks while by generative, LSTM can be used to generate similar sequences like the training samples \cite{wen2015semantically}. In this paper, we utilize the discriminative ability of LSTM for our power data classification task.

\section{Power Series Data Collection and Preliminary Analysis}
\label{stn:data_collection_ana}
In this section we present the power series data we collected followed by some preliminary analysis on the data. We will detail the simulation design rules for the data collection and the data samples collected with some pretreatment. The proposed preliminary analysis includes data visualization with different dimension reduction methods, classification results with some canonical classifiers, and feature study.
\subsection{Power Series Data Collection}
We first introduce the designing rules of the simulation for data set collection. As a data-driven study on the power series classification methodology, we need to collect a set of sample power data. The data collection should be designed carefully to make sure that the classification problem is neither trivial nor impossible to accomplish.
In this sense, our guiding line for power data collection is to collect ``different'' and ``similar'' power series:
By ``different'', the power series must be generated by different programs.
By ``similar'', the different programs can have some similar features so that the classification algorithms need to be really discriminative.

\begin{table}[]
\centering
\caption{Collected power sequences of different programs. MapReduce and web server are the two major classes, while in MapReduce there are many subclasses. In total there are 13 classes.}
\label{tbl:collected_data}
\resizebox{0.95\columnwidth}{!}{%
\begin{tabular}{ccll|c|c}
\hline
\multicolumn{4}{c|}{Program} & \multicolumn{1}{l|}{Number of sequences} & Class label \\ \hline
\multicolumn{1}{c|}{\multirow{12}{*}{MapReduce}} & \multicolumn{1}{c|}{\multirow{9}{*}{Spark}} & \multicolumn{2}{l|}{Word Count} & 100 & 0 \\ \cline{3-6}
\multicolumn{1}{c|}{} & \multicolumn{1}{c|}{} & \multicolumn{2}{l|}{Sorting} & 100 & 1 \\ \cline{3-6}
\multicolumn{1}{c|}{} & \multicolumn{1}{c|}{} & \multicolumn{2}{l|}{PI} & 100 & 2 \\ \cline{3-6}
\multicolumn{1}{c|}{} & \multicolumn{1}{c|}{} & \multicolumn{1}{c|}{\multirow{6}{*}{MLlib}} & CrossValidator & 40 & 3 \\ \cline{4-6}
\multicolumn{1}{c|}{} & \multicolumn{1}{c|}{} & \multicolumn{1}{c|}{} & Kmean & 40 & 4 \\ \cline{4-6}
\multicolumn{1}{c|}{} & \multicolumn{1}{c|}{} & \multicolumn{1}{c|}{} & LR & 40 & 5 \\ \cline{4-6}
\multicolumn{1}{c|}{} & \multicolumn{1}{c|}{} & \multicolumn{1}{c|}{} & SVM & 40 & 6 \\ \cline{4-6}
\multicolumn{1}{c|}{} & \multicolumn{1}{c|}{} & \multicolumn{1}{c|}{} & Cosine similarity & 40 & 7 \\ \cline{4-6}
\multicolumn{1}{c|}{} & \multicolumn{1}{c|}{} & \multicolumn{1}{c|}{} & PCA & 40 & 8 \\ \cline{2-6}
\multicolumn{1}{c|}{} & \multicolumn{1}{l|}{\multirow{3}{*}{Hadoop}} & \multicolumn{2}{l|}{Word Count} & 100 & 9 \\ \cline{3-6}
\multicolumn{1}{c|}{} & \multicolumn{1}{l|}{} & \multicolumn{2}{l|}{Sorting} & 100 & 10 \\ \cline{3-6}
\multicolumn{1}{c|}{} & \multicolumn{1}{l|}{} & \multicolumn{2}{l|}{PI} & 100 & 11 \\ \hline
\multicolumn{4}{l|}{Web server data} & 40 & 12 \\ \hline
\end{tabular}%
}
\end{table}

Follow the above guideline, we collected in total 13 classes of power data as shown in Table \ref{tbl:collected_data} (for convenience they are labelled as 0,1,...,12 respectively). These data fall into two major categories:
\begin{itemize}
\item Web server power data:  usually fluctuate in a continuous pattern;
\item Spark/Hadoop MapReduce programs: usually show stage-pattern, e.g. the Map stage and the Reduce stage.
\end{itemize}
For the Hadoop/Spark category, we test different programs on these platforms, some are the same for both platforms, such as the ``Word Count'' program; some only exist on one platform, for example, the ``MLlib'' programs on the Spark platform. With such design, we can achieve the proposed ``different'' and ``similar'' design goal.

Note that the collected data series are of different lengths as the running duration can vary among different programs. Although classification methods like 1NN-DTW can deal with power series of different lengths, to apply other canonical methods, in the following we cut the collected series into fixed length sequences. It is also reasonable to label sub-sequences instead of the complete power sequences of the programs as in a blind test, we have no information about the start/end point of a program. The detailed cutting method we utilize here is as shown below.

The goal is to cut the power sequences into length 200 samples. To do so, first we discard sequences with length smaller than 200 time slots (time unit: 3 seconds). The left number of sequences for each class is:
[77, 31, 30, 35, 28, 7, 40, 14, 5, 100, 58, 36, 40].
Although some power data are discarded, the total duration of the left sequences is about 199 hours and with the time unit being 3 seconds, the amount of the data are still adequate for the study.
Then these sequences are further cut into length 200 sub-sequences in the following way:
For each sequence $\mathbf{q}$ of length $n$, we cut it into multiple sequences $q[0:200]$, $q[50:250]$, .... , $q[(n-200):n]$.
We obtain 3200 test sequences in this cutting procedure, which are used as the power data in our classification study.
Note that these sequences are overlapped, as indicated by the cutting method.

Furthermore, for the purpose of multi-fold tests, we divide these samples into five folds $F_0$-$F_4$. Note that to avoid the overlapping of the training data and the test data, the fold partition is done before the sequence cutting. For each fold of test, we use $F_i$,$F_{(i+1)\%5}$ as the test data, and the left folds as the training data. 

\subsection{Preliminary Analysis}
We do some preliminary analysis on the pretreated data. The following analysis are meant to provide a basic understanding of the power data in view of classification.

\subsubsection{Basic Characteristic Analysis Based on Visualization} 
\begin{figure*}[!t]
\centering
   \includegraphics[width=1.8\columnwidth] {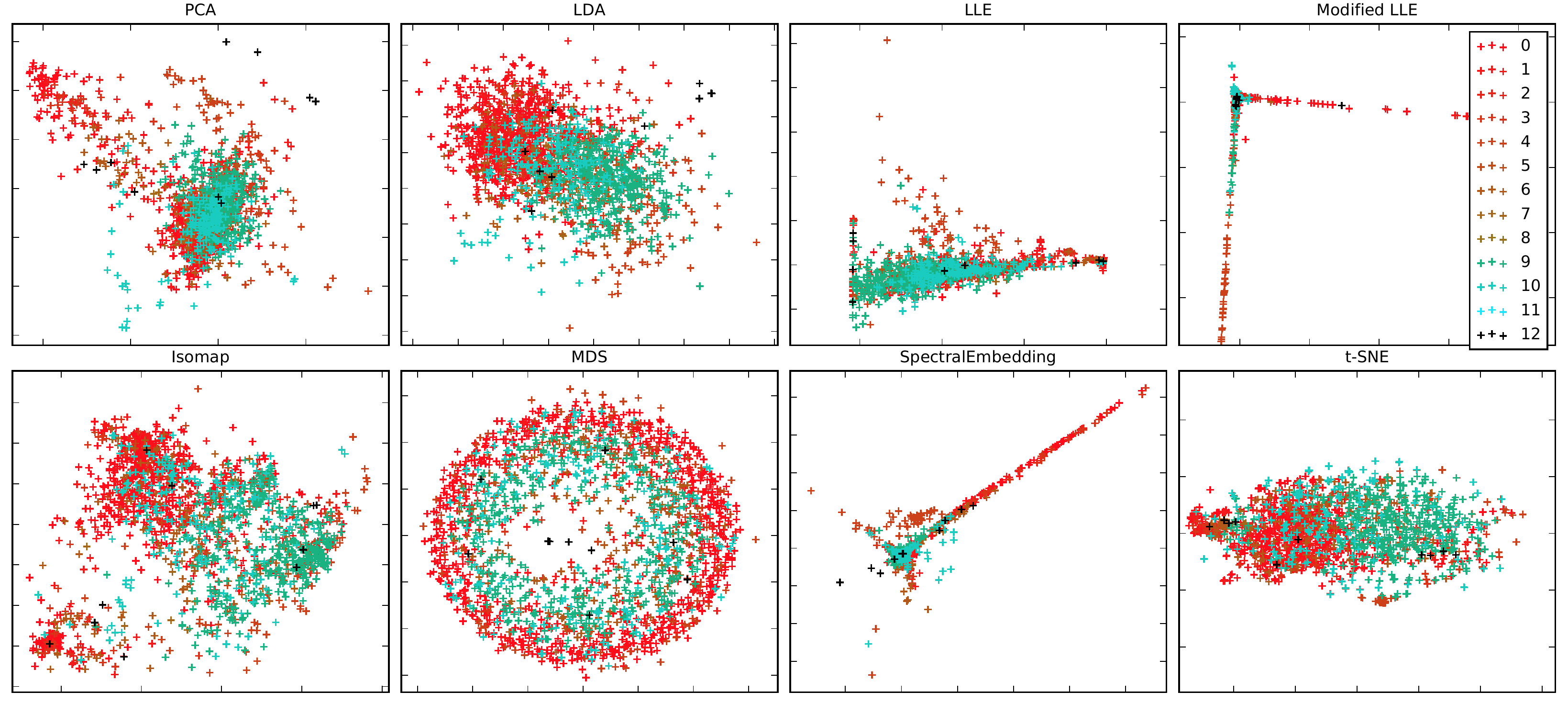}

\caption{Projection results with different manifold learning methods. The power series samples (200 dimensional data) are projected into 2-dimensional space for visualization. The results show that the power series are difficult to discriminate.}
\label{fig:manifold}
\end{figure*}

We use various dimension reduction methods to visualize the data, which can help to identify if the power series can be successfully classified to a certain degree. We utilize eight different dimension reduction methods with scikit-learn \cite{pedregosa2011scikit} and project the original fixed length power sequences into a 2-dimensional space. These dimension reduction methods are PCA, LDA, LLE, modified LLE, Isomap, MDS, Spectral Embedding and t-SNE, which are widely adopted dimension reduction methods. The 2-dimensional codes of the power data generated by these methods are shown in Fig. \ref{fig:manifold}. We use different colors to show samples from different classes.

From the Fig. \ref{fig:manifold} we can observe that the power series data are not easy to distinguish after the dimension reduction. This may be due to the short length (2 here) of the embedding code; however, it still shows that the power series classification task cannot be easily done.

\subsubsection{Tackling the Classification Problem with the Canonical Classifiers} We test some canonical classifiers to tackle the power series classification problem. The canonical classifiers tested here are listed as follows: Nearest Neighbours, Linear SVM, RBF SVM, Decision Tree, Random Forest, AdaBoost, Naive Bayes, LDA and QDA \cite{pedregosa2011scikit} . Parameter settings for these classifiers are tuned manually. The classification results of these methods are shown in Table \ref{tbl:results_cano}.

\begin{table*}[t]
\centering
\caption{The classification results of the canonical classifiers. These classifiers can achieve accuracy up to about 60\%.}
\label{tbl:results_cano}
\resizebox{1.95\columnwidth}{!}{%
\begin{tabular}{c|c|c|c|c|c|c|c|c|c}
\hline
Test case & Nearest Neighbours & Linear SVM & RBF SVM & Decision Tree & Random Forest & AdaBoost & Naive Bayes & LDA & QDA \\ \hline
Fold 0 & 0.4346 & 0.5187 & 0.3175 & 0.5747 & 0.5891 & 0.4593 & 0.4406 & 0.4380 & 0.4092 \\ \hline
Fold 1 & 0.4427 & 0.5105 & 0.3063 & 0.5607 & 0.6050 & 0.4192 & 0.4360 & 0.4377 & 0.4276 \\ \hline
Fold 2 & 0.4420 & 0.5071 & 0.3283 & 0.5901 & 0.6133 & 0.4121 & 0.4016 & 0.4256 & 0.4248 \\ \hline
Fold 3 & 0.4475 & 0.5004 & 0.3094 & 0.5835 & 0.6378 & 0.3813 & 0.4123 & 0.3707 & 0.4574 \\ \hline
Fold 4 & 0.4673 & 0.5383 & 0.3301 & 0.5610 & 0.6174 & 0.3834 & 0.4278 & 0.4407 & 0.4786 \\ \hline
\end{tabular}%
}
\end{table*}

From the results we can observe that for a 13-classes classification problem, the highest accuracy achieved by these methods are about 60\% (by Random forest). The classification accuracy is not promising (when compared to the 1NN-DTW shown below), which actually proves that our power series labelling problem is a typical time series classification problem, as stated in \cite{xi2006fast}, for such problem, canonical Euclidean distance metric based classifiers cannot achieve good results usually.

\subsubsection{Feature Based Classification Study} In general, as a signal classification problem, the power series labelling problem can be solved by first extracting certain features from the raw power series and then carry out the classification with these features. In this subsection, we study such possibility and test power series classification with the DFT \cite{burrus1991dft} feature of the original power sequences. With DFT, each power sequence can be transformed into the spectrum space resulting a new representation. The spectrum representation can be aligned as a vector as the input to the classifiers. We test the classification result of 1NN-DTW with the raw data compared to with the DFT feature. The classification results are shown in Table \ref{tbl:results_DFT}. Note that for the 1NN-DTW, the maximum offset $r$ is set to 15\% of the sample length, which is manually tuned in the experiment.

\begin{table}[]
\centering
\caption{Classification results of the 1NN-DTW with the original series and with the DFT feature.}
\label{tbl:results_DFT}
\resizebox{0.7\columnwidth}{!}{%
\begin{tabular}{c|c|c}
\hline
Test case & \begin{tabular}[c]{@{}c@{}}1NN-DTW with\\original series\end{tabular} & \begin{tabular}[c]{@{}c@{}}1NN-DTW with\\DFT feature\end{tabular} \\ \hline
Fold 0 & 0.8548 & 0.7122 \\ \hline
Fold 1 & 0.8393 & 0.7029 \\ \hline
Fold 2 & 0.8369 & 0.6761 \\ \hline
Fold 3 & 0.8329 & 0.6998 \\ \hline
Fold 4 & 0.8514 & 0.6885 \\ \hline
\end{tabular}%
}
\end{table}

From Table \ref{tbl:results_DFT} we can observe that the DFT features are not helping. The reason is that classification with the original data can maximize the information used in classification, while the DFT feature is less informative.

To summary, we find that the power series classification problem is not easy to tackle especially with the canonical classifiers and with some common used features. In the following, we will propose a new distance measurement inspired from DTW and combine it with the state-of-art sequence modelling neural network LSTM.

\section{The Proposed Power Series Classification Algorithm}
\label{stn:new_algorithm}

In this section we present the proposed new power series classification algorithm which hybridizes a nearest neighbour classifier with a novel distance measurement and a LSTM classifier. In the following we first introduce the two components respectively and then present the hybrid algorithm.

\subsection{Nearest Neighbour with the Local Time Warping (LTW)}
We propose a new classifier which utilize a novel distance measurement to compute the distance between two sequences which we termed as Local Time Warping (LTW) as its warping computation for each time step is done in a local window without a dynamic programming procedure like DTW. The LTW is developed to replace the DTW distance measurement in the 1NN-DTW classifier. 

The idea behind LTW is as follows. Comparing the algorithms of DTW and the Euclidean distance, the major difference in between is that there are a lot of ``min'' operators in DTW. Such ``min'' operator actually is the key to the ``warping'' map between the two time series. Despite the warping operation, DTW utilizes a dynamic programming procedure to optimize the mapping. Note that dynamic programming is slow and it is not directly optimizing the classification accuracy. In this case, what if we do not use the dynamic programming procedure? We may try some low cost warping operations; is it possible that such a distance measurement can be as good as DTW? Here we propose the LTW to answer this question. Also, note that the DTW is computed by a beautiful symmetric formula which makes it commutative for the two time series in computing the distance. What if we do not need the distance measurement to be commutative? Can it be better with the non-commutative feature? Our proposed LTW will also answer this problem. The detailed design is shown below. 

The LTW distance measurement is computed in the following way. Suppose we have two sequences $\mathbf{x}$ and $\mathbf{y}$, both of length $n$. We define the LTW distance between $\mathbf{x}$ and $\mathbf{y}$ as:
\begin{equation}
d_{k}^{LTW}(\mathbf{x},\mathbf{y})=\sum_{i=k+1}^{n-k}\min(|x_{i}-y_{i}|,|x_{i}-y_{i+1}|,|x_{i}-y_{i+k}|), \label{eq_metaltw}
\end{equation}
\begin{equation}
LTW_{G}(\mathbf{x},\mathbf{y})=\sum_{k\in G}d{}_{k}^{LTW}(x,y). \label{eq_ltw}
\end{equation}

As shown in \eqref{eq_ltw}, LTW works in the following manner. In computing the distance between $\mathbf{x}$ and $\mathbf{y}$ (when we want to find a nearest neighbour of $\mathbf{x}$), we set $\mathbf{x}$ as the base sequence and test the similarity of $\mathbf{y}$ to $\mathbf{x}$ in the following way: with a warping index set $G$, for $k\in G$, for time step $i$ in $\mathbf{x}$, we compute the minimum absolute distance between $x_i$ and one of $y_i$, $y_{i+1}$, $y_{i+k}$; then we add these distance measures for $i=k+1,...,n-k$ and for $k\in G$ up, which is the LTW distance from $\mathbf{y}$ to $\mathbf{x}$ with warping index $G$. Note that \eqref{eq_metaltw} is a linear algorithm (only three items to compare no matter how large $k$ is).

Note that the $LTW_G(\mathbf{x},\mathbf{y})$ distance is non-commutative, which means that $LTW_G(\mathbf{x},\mathbf{y}) \neq LTW_G(\mathbf{y},\mathbf{x})$ can be true. We use $LTW_G(\mathbf{x},\mathbf{y})$ to compute the nearest neighbour of sequence $\mathbf{x}$, in a sense that to find the best match of $\mathbf{x}$ among the other samples such as $\mathbf{y}$. For comparison, the DTW distance is obviously commutative. The non-commutative feature of LTW can be beneficial, as our target is to find the nearest neighbour for each $\mathbf{x}$. A non-commutative distance measurement is enough to serve the purpose and can provide more flexibility by enforcing less constraints to the distance measurement.

\subsection{Long Short Term Memory Neural Network}
We utilize the LSTM classifier following \cite{LSTM} for our power series classification problem. The LSTM neural network consists of an input layer, a LSTM layer and a logistic regression layer as depicted in Fig. \ref{fig:LSTM}. The three layers function in the following way respectively:

\begin{figure}[!t]
\centering
   \includegraphics[width=0.95\columnwidth] {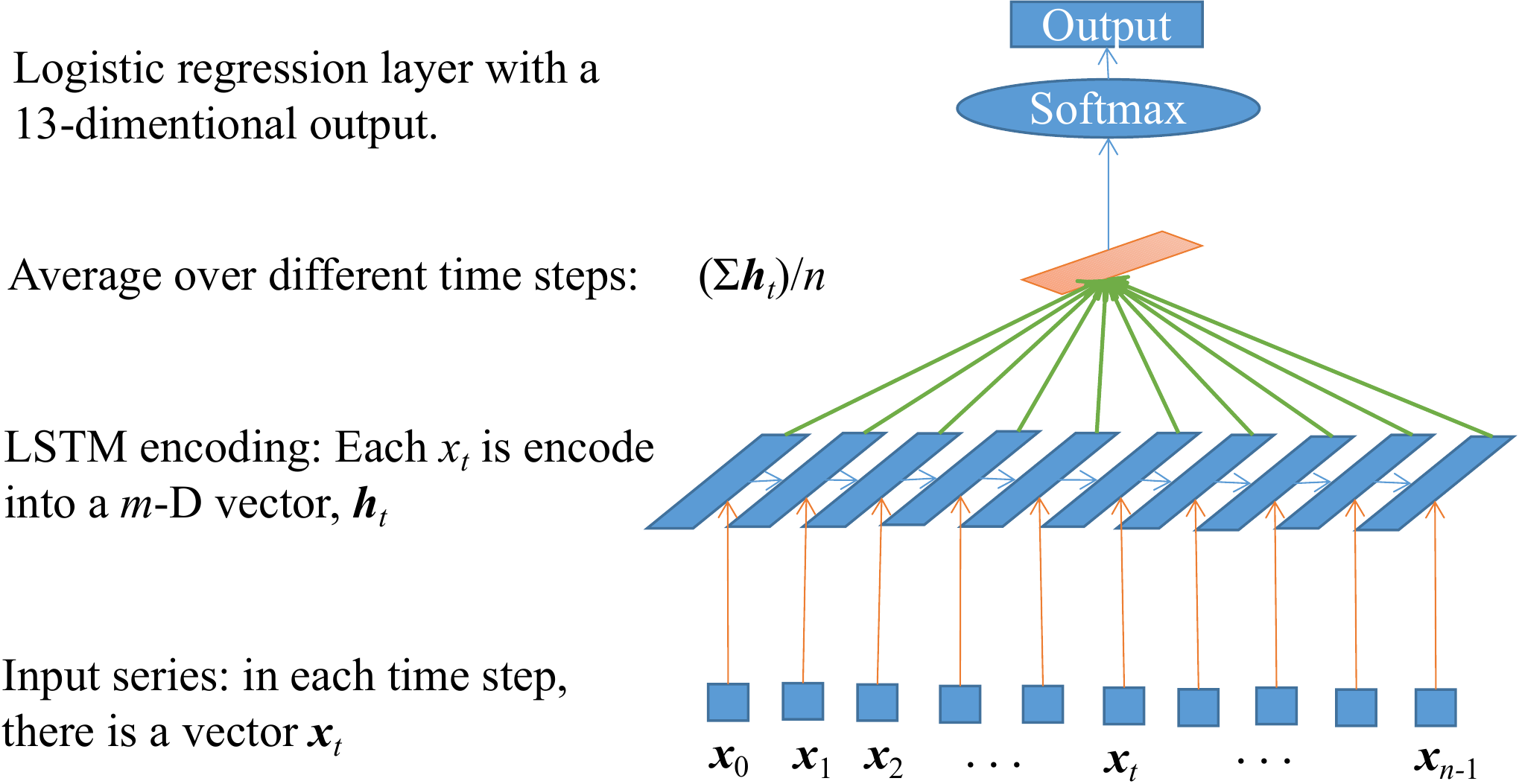}

\caption{Illustration of the LSTM neural network. It contains three layers: the input layer, the LSTM layer and the logistic regression layer. The output is a 13-dimensional vector which denotes the probability of the sample belonging to each class.}
\label{fig:LSTM}
\end{figure}

\begin{itemize}
\item Input layer: The input data sample, which is a length $n$ vector $\mathbf{x}$, is firstly discretized into range [0, $S$]. Such an operation is a smoothing operation to the original power series, which can affect the performance of the LSTM. Then each time step $x_t$, $t=0,...,n-1$ is enriched into a $m$-dimensional vector $\mathbf{x}_t$ which can ease the following computation, i.e. $\mathbf{x}_t=x_t \cdot \mathbf{e}$, where $\mathbf{e}$ is a $m$-dimensional vector with all entries equal to 1. After the above process, the new sequence $\mathbf{x}_0, \mathbf{x}_1,...,\mathbf{x}_{n-1}$ is used as the input to the LSTM layer.

\item LSTM layer: The LSTM layer contains $n$ LSTM node, where each LSTM node $t$ can output an $m$ dimensional code $\mathbf{h}_t$. The operation inside the LSTM node is shown below. First, for each time step $\mathbf{x}_t$, the LSTM node needs to compute a new state denoted by $\mathbf{C}_t$. To compute $\mathbf{C}_t$, a candidate state $\mathbf{C}'_t$ is firstly computed as:
\begin{equation}
\mathbf{C}'_t=tanh(\mathbf{W}_c\mathbf{x}_t+\mathbf{U}_c\mathbf{h}_{t-1}+\mathbf{b}_c).
\end{equation}
Then two gates, an input gate $i_t$ and a forget gate $f_t$ are computed to update the new state:
\begin{equation}
i_t=\sigma(\mathbf{W}_i\mathbf{x}_t+\mathbf{U}_i\mathbf{h}_{t-1}+\mathbf{b}_i).
\end{equation}
\begin{equation}
f_t=\sigma (\mathbf{W}_f \mathbf{x}_t+\mathbf{U}_f \mathbf{h}_{t-1}+ \mathbf{b}_f).
\end{equation}
Then the new state of the LSTM node is computed as:
\begin{equation}
\mathbf{C}_t=i_t \cdot \mathbf{C}_t + f_t \cdot \mathbf{C}'_t.
\end{equation}

With the node state, to further compute the output, an output gate is firstly computed as:
\begin{equation}
o_t=\sigma(\mathbf{W}_o\mathbf{x}_t+\mathbf{U}_o\mathbf{h}_{t-1}+\mathbf{b}_o).
\end{equation}
Finally the output of a LSTM node $t$ is computed as:
\begin{equation}
\mathbf{h}_t = o_t \cdot tanh(\mathbf{C}_t).
\end{equation}

The output of all LSTM nodes are then added together as the output of the LSTM layer:
\begin{equation}
\mathbf{h}=\sum_{t=0}^{n-1}\mathbf{h}_t.
\end{equation}

\item Logistic regression layer: In this layer the output of the LSTM layer is used to compute the label of the test sample in the following way. First we use the $softmax$ \cite{hosmer2004applied} function to compute the probability vector $\mathbf{P}$ with its each entry representing the probability of the test sample belonging to a class:
\begin{equation}
\mathbf{P}=softmax(W h +b). \label{eq_probv}
\end{equation}
Then the prediction $y_{pred}$ is the class which achieves the largest probability:
\begin{equation}
y_{pred}=argmax_i(\mathbf{P}).
\end{equation}

\end{itemize}

To train the LSTM classifier, the loss function is defined as the negative log-likelihood function with the label of the training data $\mathbf{y}$:
\begin{equation}
-L(\theta, D)=-\sum_{i=0}^{|D|-1}log(P_{y^{(i)}}|\mathbf{x}^{(i)},\theta)),
\end{equation}
where $\theta$ is the set of all the weight and bias parameters in the LSTM neural network (which are adjusted in the training process); $D$ is a batch of training samples. Size of $D$ can be important for the performance of of the classifier.

\subsection{Hybridization of LSTM and 1NN-LTW}
In this subsection we propose to combine the 1NN-LTW classifier and the LSTM classifier. The underlying rational is that both classifier can achieve high classification accuracy for our problem but in very different manners: the 1NN-LTW is a nearest neighbour classifier, which is a data-based classifier without a training process; while LSTM is a training based classifier in which the training data are firstly used to build a model and then the model is used to classify the test data. In our experiments, both classifiers can perform promisingly individually; however, our numerical simulation shows that the accurately classified samples by the two classifiers have significant differences. In such sense, we propose to combine the two algorithms to construct a even stronger classifier.

The hybrid algorithm is designed in the following way. Considering that in practice, the training of LSTM and the computing of the distance matrix for 1NN-LTW can be both time consuming, the hybrid algorithm is designed to be as simple as possible. We first obtain the two individual classifiers $C_{LTW}$ (the nearest neighbour classifier with CID enhanced LTW) and $C_{LSTM}$ (the trained LSTM classifier). For each classifier, we obtain the probability vector when classifying some test time series $x$: $\mathbf{p}_{LTW}(\mathbf{x})$ and $\mathbf{p}_{LSTM}(\mathbf{x})$, where $\mathbf{p}_{LTW}(\mathbf{x})$ is defined as:
\begin{equation}
\mathbf{p}_{LTW}(\mathbf{x})=\frac{\sum_{i=1}^{m}(m-i)\mathbf{v}_i}{\Delta},
\end{equation}
where $\mathbf{v}_i$ is a all zero vector except its value at the index of the class of the $i^th$ neighbour obtained from $C_{LTW}$ is to be 1. $\Delta$ is a normalization vector which is used to make sure that the summation of the obtained probability vector equal to 1. $\mathbf{p}_{LSTM}(x)$ is obtained equation \eqref{eq_probv}. 

With this two probability vector, we simply add them up and the test series will be classified to be the class index with the maximum probability, i.e.:
\begin{equation}
C_{LSTM/LTW}(\mathbf{x})=argmax(\mathbf{p}_{LTW}(\mathbf{x})+\mathbf{p}_{LSTM}(\mathbf{x})). \label{eq_hybridc}
\end{equation}

The detailed algorithm is shown in Algorithm \ref{alg:training}.

\begin{algorithm}[!htb]
\caption{Hybridization of 1NN-LTW and LSTM}
\label{alg:training}
\begin{algorithmic}[1]
    \STATE  Individual training process: build classifier $C_{LTW}$ and $C_LSTM$. Train $C_LSTM$ with the training data.
    \STATE  For test time series $\mathbf{x}$, compute the probability vector $\mathbf{p}_{LTW}(\mathbf{x})$ and $\mathbf{p}_{LSTM}(\mathbf{x})$.
    \STATE  Classify the test time series $\mathbf{x}$ according to \eqref{eq_hybridc} and get the label $l_{hybrid}(\mathbf{x})$.
    \STATE \emph{Output}: $l_{hybrid}(\mathbf{x})$, as the predicted label for sample $\mathbf{x}$.
\end{algorithmic}
\end{algorithm}

\section{Numerical Evaluation and Analysis}
\label{stn:Numer}
In this section we present the experimental results of the above proposed algorithms and the analysis. We first conduct experiments to investigate the proposed LTW and compraed it with various variants of DTW. Then we compare the classification accuracy of the proposed 1NN-LTW, LSTM and their hybrid algorithm LSTM/LTW with the baseline algorithm 1NN-DTW. For the base line algorithm 1NN-DTW, the maximum warping offset $w$ is manually tuned and set to be 30. For the 1NN-LTW, the warping index set $G$ of the LTW distance is set to $\{1,2,...,10\}$, and we will show analysis on the affects of the set $G$. For the LSTM neural network, we set the maximum number of epoch to be 50. For some key parameters which can affect the performance of LSTM ,we give a detailed discussion in the following parameter settings study. Test data and codes for the LTW tests are available at https://www.dropbox.com/s/lylsece6xysayw8/DataAndCode.zi p?dl=0. In presenting the classification results, for convenience, we will simply use test Fold $i$ to denote the test with test samples in $F_i$ and $F_{(i+1)\%5}$.

\subsection{Experimental Study on LTW}
In this subsection, we conduct experiments to prove that the proposed LTW is indeed a different distance measurement from the existed DTW variants, and prove that it works better or nearly the same(for the linear version) to DTW and its various state-of-art variants. We will also prove that the non-communicative feature is indeed beneficial. Note that we will not try to use massive experimental data to prove that the proposed LTW is superior than other DTW variants, which is indeed not true not only because of the No Free Launch theory, but also because these distance measurements are mostly designed in a way without a training objective to directly optimize the classification accuracy: for example, in DTW, the dynamic programming process can optimize the match; however, such optimization goal is different from the classification accuracy. In such sense, all these distance measurements can suffer from model bias and when they are applied to different data sets, their performance will definitely vary. As proved in \cite{bagnall2016great}, only ensemble based methods can significantly outperform 1NN-DTW by more than 3\%. In this section, we will only compare LTW with the DTW with Manhattan distance (DTWm), DTW with Euclidean distance(we will use this as the default DTW in this paper, as it has better performance), and DTW variants MSM, LB\_Keogh, and the enhancer CID as these methods performs good in general as shown in \cite{bagnall2016great}. Details are shown below.

First, we present the comparison of 1NN-LTW ($G=\{1,2...,10\}$) with 1NN-DTWm (warping window $w=30$), 1NN-DTW (warping window $w=30$), MSM (with its threshold parameter $c=1.0$). The results on our data set are shown in Table \ref{tbl:results_LTW1}. Parameter settings for 1NN-DTW and MSM are tuned by manually. Clearly the performance of 1NN-LTW is better.

Second, we compare the fast linear LTW with $G=\{10\}$ with the linear runtime lower bound method LB\_Keogh. We test LB\_Keogh with window size $w=5$ and $w=10$ respectively. The results are shown in Table \ref{tbl:lbk} and LTW with $G=\{10\}$ outperforms LB\_Keogh significantly on our data set. This proves that the fast linear version of LTW can still perform quite good.  

Third, we present the classification with the CID enhanced distance measurement. We show the result of CID(DTW) and CID(LTW) with $G=\{1,2...,10\}$ and $G=\{10\}$ in Table \ref{tbl:cid}. Clearly, the CID can improve the performance of both DTW and LTW for our data set. With CID modifier, the LTW is still slightly better than DTW; however, the advantage of CID(LTW) over CID(DTW) becomes smaller, we believe that it is reasonable as the accuracy is upper bounded and it will be much more difficult to further improve the accuracy when the algorithm is already very accurate. 

At last, we present the experimental results to show that the non-commutative feature of LTW is indeed beneficial. To do so, we implement a simple commutative version of LTW defined as:
\begin{equation}
LTW^{com}(\mathbf{x},\mathbf{y})=LTW(\mathbf{x},\mathbf{y})+LTW(\mathbf{y},\mathbf{x}).
\end{equation}
The experimental results compare the LTW with the $LTW^{com}$ is shown in Table \ref{tbl:com}. Clearly, LTW outperforms its commutative version significantly. This proves that the non-commutative feature of LTW is indeed beneficial.

\begin{table}[]
\centering
\caption{Classification accuracy of 1NN-LTW ($G=\{1,2...,10\}$) with 1NN-DTWm ($w=30$), 1NN-DTW ($w=30$), MSM ($c=1.0$).}
\label{tbl:results_LTW1}
\resizebox{0.95\columnwidth}{!}{%
\begin{tabular}{c|c|c|c|c}
\hline
Test case & 1NN-LTW(\{1,...,10\}) & 1NN-DTWm & 1NN-DTW & MSM \\ \hline
Fold 0 & 0.8862 & 0.7988 & 0.8480 & 0.8294 \\ \hline
Fold 1 & 0.8854 & 0.7958 & 0.8351 & 0.8310 \\ \hline
Fold 2 & 0.8661 & 0.8220 & 0.8347 & 0.8699 \\ \hline
Fold 3 & 0.8809 & 0.8118 & 0.8379 & 0.8682 \\ \hline
Fold 4 & 0.8894 & 0.8168 & 0.8442 & 0.8515 \\ \hline
\end{tabular}%
}
\end{table}

\begin{table}[]
\centering
\caption{Comparison of 1NN-LTW ($G=\{10\}$) with 1NN-LB\_Keogh ($w=5,10$).}
\label{tbl:lbk}
\resizebox{0.8\columnwidth}{!}{%
\begin{tabular}{c|c|c|c}
\hline
 & 1NN-LTW($G=\{10\}$) & LB\_Keogh ($w=5$) & LB\_Keogh ($w=10$) \\ \hline
Fold 0 & 0.8829 & 0.5866 & 0.4542 \\ \hline
Fold 1 & 0.8762 & 0.5791 & 0.4753 \\ \hline
Fold 2 & 0.8444 & 0.6081 & 0.4592 \\ \hline
Fold 3 & 0.8696 & 0.6018 & 0.4616 \\ \hline
Fold 4 & 0.8838 & 0.6053 & 0.4326 \\ \hline
\end{tabular}%
}
\end{table}

\begin{table}[]
\centering
\caption{Classification results with CID enhanced distance measurement.}
\label{tbl:cid}
\resizebox{0.95\columnwidth}{!}{%
\begin{tabular}{c|c|c|c}
\hline
 & 1NN-CID(DTW) & 1NN-CID(LTW\{1,...,10\}) & 1NN-CID(LTW\{10\}) \\ \hline
Fold 0 & 0.8829 & 0.9041 & 0.8837 \\ \hline
Fold 1 & 0.8854 & 0.8904 & 0.8912 \\ \hline
Fold 2 & 0.8833 & 0.8684 & 0.8579 \\ \hline
Fold 3 & 0.8950 & 0.9013 & 0.8887 \\ \hline
Fold 4 & 0.8999 & 0.9031 & 0.8846 \\ \hline
\end{tabular}%
}
\end{table}

\begin{table}[]
\centering
\caption{Comparison of LTW and the $LTW^{com}$ }
\label{tbl:com}
\resizebox{0.5\columnwidth}{!}{%
\begin{tabular}{c|c|c}
\hline
 & 1NN-LTW & 1NN-$LTW^{com}$ \\ \hline
Fold 0 & 0.8862 & 0.7385 \\ \hline
Fold 1 & 0.8853 & 0.7029 \\ \hline
Fold 2 & 0.8661 & 0.7218 \\ \hline
Fold 3 & 0.8809 & 0.7294 \\ \hline
Fold 4 & 0.8894 & 0.7692 \\ \hline
\end{tabular}%
}
\end{table}

\subsection{The Classification Accuracy Rate Comparison}
The five-fold classification accuracy results for the hybrid algorithm LSTM/LTW are shown in Table \ref{tbl:results_LSTM_LTW},compared with the non-hybrid classifiers. Results of LSTM and LSTM/LTW are based on 10 independent runs. 
\begin{table}[]
\centering
\caption{Classification accuracy of 1NN-LTW,LSTM and the hybrid algorithm LSTM/LTW compared to the baseline algorithm 1NN-DTW.}
\label{tbl:results_LSTM_LTW}
\resizebox{0.95\columnwidth}{!}{%
\begin{tabular}{c|c|c|c|c}
\hline
Test case & 1NN-DTW & 1NN-CID(LTW) & LSTM & LSTM/LTW \\ \hline
Fold 0 & 0.8548 & 0.9040 & 0.8772$\pm$0.0153 & 0.9267$\pm$0.0030 \\ \hline
Fold 1 & 0.8393 & 0.8903 & 0.8780$\pm$0.0130 & 0.9115$\pm$0.0046 \\ \hline
Fold 2 & 0.8369 & 0.8683 & 0.8547$\pm$0.0171 & 0.8923$\pm$0.0038 \\ \hline
Fold 3 & 0.8329 & 0.9013 & 0.8772$\pm$0.0077 & 0.9167$\pm$0.0021 \\ \hline
Fold 4 & 0.8514 & 0.9031 & 0.8778$\pm$0.0061 & 0.9256$\pm$0.0051 \\ \hline
\end{tabular}%
}
\end{table}

From Table \ref{tbl:results_LSTM_LTW} we can observe that:
\begin{itemize}
\item The proposed LSTM classifier shows similar accuracy compared to 1NN-LTW and it also outperforms 1NN-DTW on our data set.
\item The hybrid algorithm LSTM/LTW can achieve higher accuracy compared to 1NN-LTW and LSTM by an increment of about 3\%, which proves that the hybrid algorithm can indeed improve the classification accuracy.
\end{itemize}

For the first observation, we can see that LSTM, as a neural network, can significantly outperform the other canonical classifiers like SVM, which proves its strong modelling ability for sequential data. Note that a common neural network like multilayer perceptron (MLP) can not perform as good as LSTM. The performance of LSTM can be seriously affected by the training settings, which we will discuss below.

For the second observation, we can see that the improvement is small, which is reasonable as the baseline algorithms already achieve a high accuracy individually, making it difficult to achieve large improvement for the hybrid algorithm. The improvement caused by the hybrid algorithm will be shown clearly in the following detailed analysis.

\subsection{Analysis on the Accurately Classified Samples}
In this subsection we analyse the accurately classified samples of the power series and study the difference between different classifiers. In doing so we will be able to identify why and how the hybrid algorithm works.

\begin{figure*}[!t]
\centering
   \includegraphics[width=1.95\columnwidth] {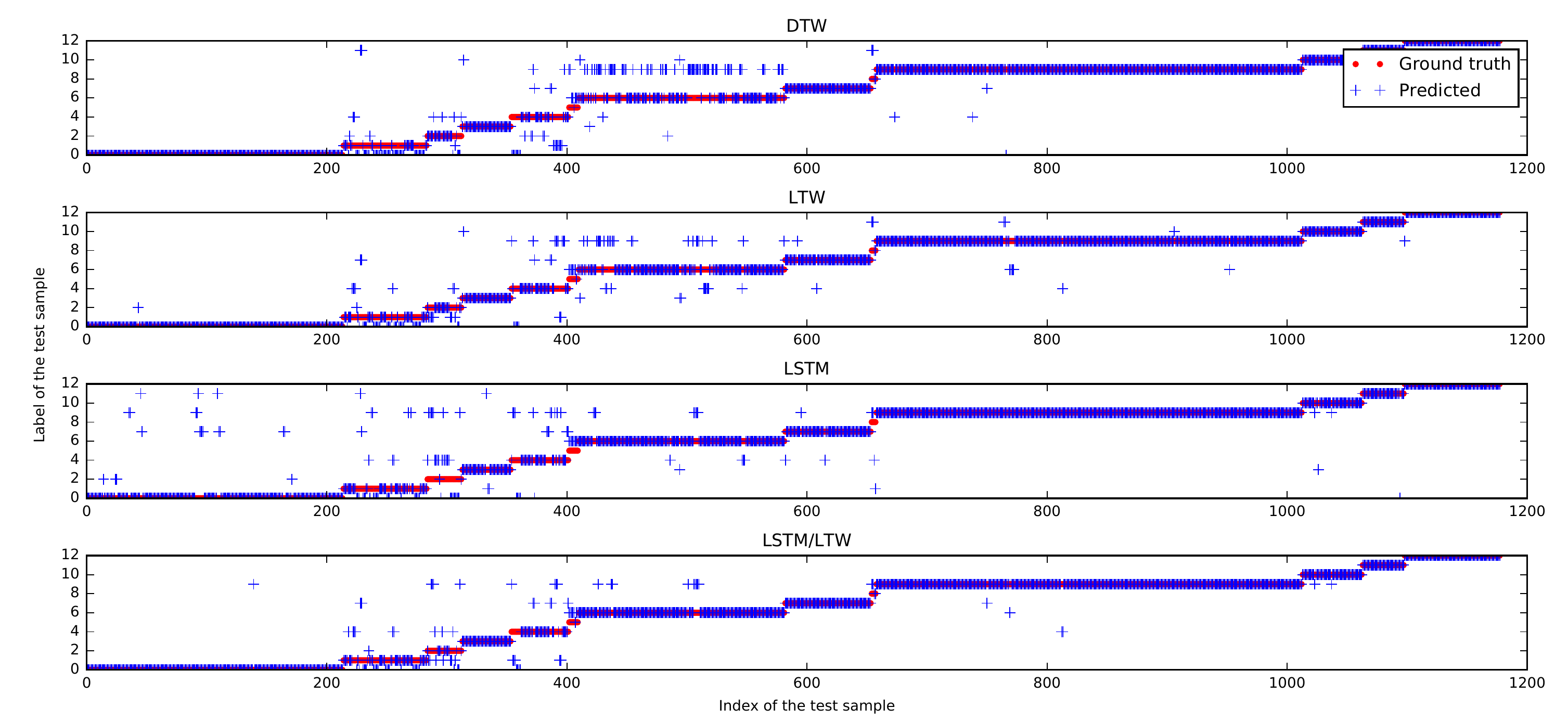}

\caption{Classification response for the different algorithms on Fold 0. The ground truth/predicted labels of the test samples are plotted against the index of the sample. Clearly the hybrid algorithm gains the advantage of both LSTM and 1NN-LTW.}
\label{fig:response}
\end{figure*}

\begin{figure*}[!t]
\centering
\subfigure[]{
   \includegraphics[width=1.95\columnwidth] {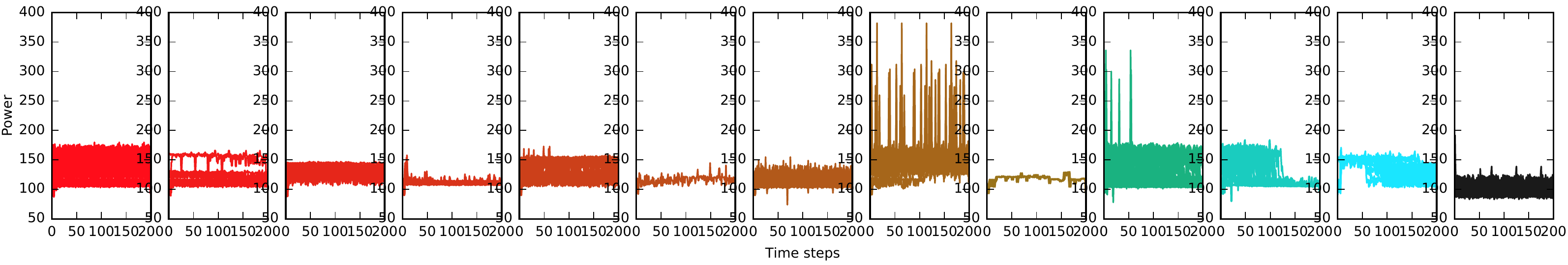}
}
\subfigure[]{
   \includegraphics[width=1.95\columnwidth] {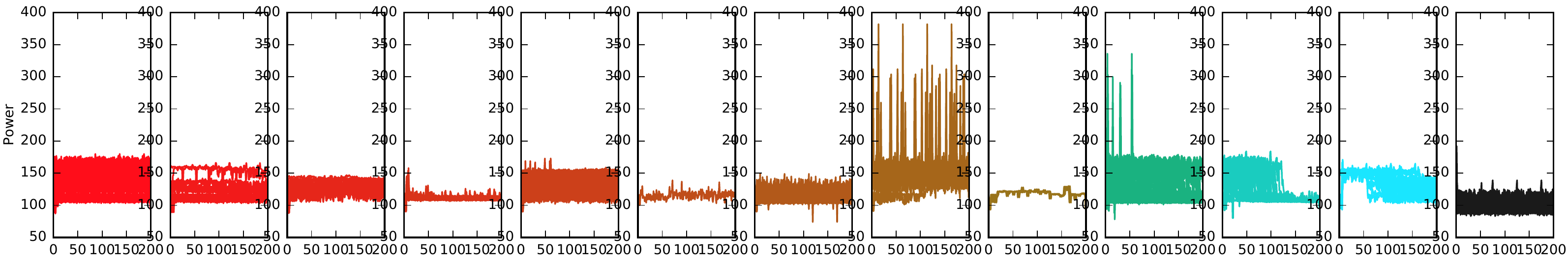}
}
\subfigure[]{
   \includegraphics[width=1.95\columnwidth] {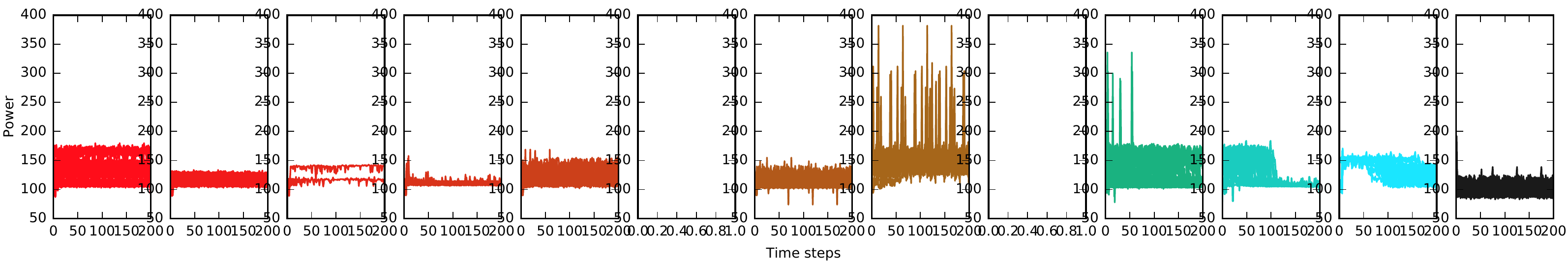}
}
\subfigure[]{
   \includegraphics[width=1.95\columnwidth] {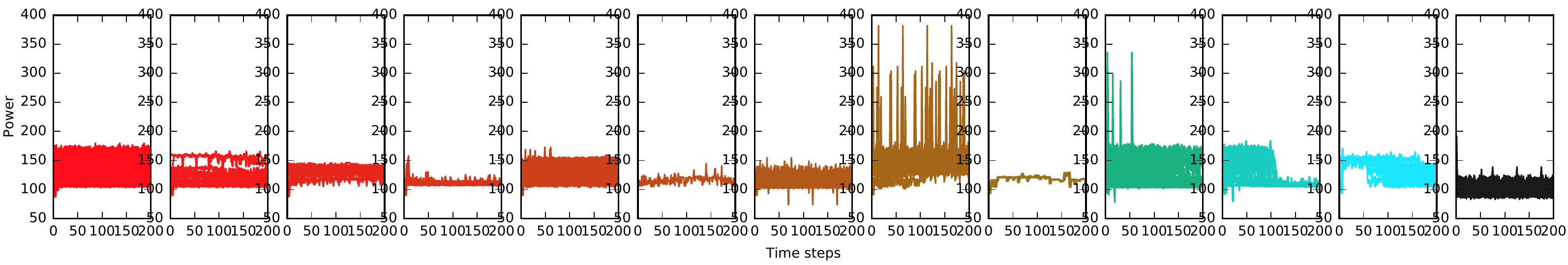}
}

\caption{Accurately classified samples for the different algorithms on Fold 0. Samples from the same class are drawn in the same sub-figure: (a)1NN-DTW; (b) 1NN-LTW; (c) LSTM; (d) LSTM/LTW. LSTM cannot classify some classes while the hybrid algorithm gains the advantage of both LSTM and 1NN-LTW.}
\label{fig:acc_samples}
\end{figure*}

Fig. \ref{fig:response} shows the predicted labels for the test samples in Fold 0 of the 1NN-DTW, 1NN-LTW, LSTM and the hybrid algorithm LSTM/LTW. Fig. \ref{fig:acc_samples} shows the accurately classified samples for each class and for each algorithm. From Fig. \ref{fig:response} and \ref{fig:acc_samples} we can observe that:
\begin{itemize}
\item The proposed 1NN-LTW method performs similarly to 1NN-DTW, although 1NN-LTW can accurately predict more test samples. This is reasonable as the two classifiers are both nearest-neighbour classifiers and they have similar measurement definition.
\item The proposed LSTM classifier shows certain degree of difference compared to the other two nearest neighbour classifiers. One example, the LSTM classifier cannot predict any test samples from the Spark-MLlib-LR (class label 5) and the Spark-MLlib-PCA (class label 8) classes, while both 1NN-DTW and 1NN-LTW can; however, LSTM performs better than the other algorithms on classes Spark-MLlib-SVM (class label 6) and Hadoop-WordCount (class label 9).
\item The proposed LSTM/LTW classifier can successfully combine the advantages of LSTM and 1NN-LTW. Such as for the Spark-MLlib-LR class and Hadoop-WordCount classes, the hybrid algorithm achieve similar performance to the better one of LSTM and 1NN-LTW.  
\item All the classifiers can successfully classify the test samples of the web server class, which is reasonable as the web server program is of a completely different kind from the other MapReduce programs.
\end{itemize}

The above results show the difference of the 1NN-LTW and the LSTM classifier which makes the hybrid algorithm work. Although 1NN-LTW and the LSTM can achieve similar accuracy, their accurately classified samples have significant differences. To make this more clearly, we compute the union-accuracy $acc_{union}$ of the two classifiers as follows:
\begin{equation}
acc_{union}=\frac{|A_{LSTM} \cup A_{1NN-LTW}|}{N},
\end{equation}
where $A_{LSTM}$ and $A_{1NN-LTW}$ are the sets of the accurately classified samples by LSTM and 1NN-LTW respectively; $N$ is the total number of test samples in this test fold.
The union-accuracy of the five fold tests are shown in Table \ref{tbl:union_acc}. It can be seen that the union-accuracy is between 94\%-96\%. It shows the potentiality of a hybrid algorithm of the two classifiers. Note that the hybrid algorithm can only achieve accuracy smaller than the union-accuracy, as the union-accuracy is computed in an ideal manner. 

\begin{table}[]
\centering
\caption{The union-accuracy of the 1NN-LTW and the LSTM classifiers in the five fold tests.}
\label{tbl:union_acc}
\resizebox{0.4\columnwidth}{!}{%
\begin{tabular}{c|c}
\hline
Test case & $acc_{union}$ \\ \hline
Fold 0 & 0.9482 \\ \hline
Fold 1 & 0.9489 \\ \hline
Fold 2 & 0.9468 \\ \hline
Fold 3 & 0.9541 \\ \hline
Fold 4 & 0.9612 \\ \hline
\end{tabular}%
}
\end{table}

\subsection{Discussion on the Parameter Settings}
In this subsection we discuss the parameters settings in the above algorithms. First we study the parameter used in the LTW measurement, the warping index set $G$. The test results with different $G$ settings are shown in Table \ref{tbl:LTW_order}. It can be seen that a proper $G$ setting is needed as a set G too small can deteriorate the performance. In our experiments we find that with a larger set $G$, the performance of LTW is more stable. Note that increasing the size of $G$ can cause higher computing cost.

\begin{table}[]
\centering
\caption{Test results with different settings for the warping index set $G$ in 1NN-LTW}
\label{tbl:LTW_order}
\resizebox{0.95\columnwidth}{!}{%
\begin{tabular}{|c|c|c|c|c|}
\hline
 & $G=\{1\}$ & $G=\{1,..,4\}$ & $G=\{1,...,8\}$ & $G=\{1,...,12\}$ \\ \hline
Fold 0 & 0.8166 & 0.8591 & 0.8727 & 0.8846 \\ \hline
Fold 1 & 0.8075 & 0.8552 & 0.8728 & 0.8904 \\ \hline
Fold 2 & 0.8138 & 0.8601 & 0.8684 & 0.8661 \\ \hline
Fold 3 & 0.8132 & 0.8668 & 0.8760 & 0.8802 \\ \hline
Fold 4 & 0.8103 & 0.8571 & 0.8878 & 0.8902 \\ \hline
\end{tabular}%
}
\end{table}

Second, we discuss the parameter settings for the LSTM classifier. Tuning of the hyper-parameters of the LSTM network is critical. In our experiments, we find that an improper setting can result a bad performance with accuracy lower than 50\% for the LSTM. We find the following key settings in the LSTM classifier, which we have tested and find the proper setting, although detailed experimental results are omitted here. The hyper-parameter settings: three parameters are specially tuned in our experiments, which are the batch size (we set to 60), the dimension of the LSTM node (we set to 90), and the discretized range parameter $S$ (we set to 100). We also tested two more different implementation variations of LSTM: 1)Adding a dropout layer, which is tested and not helpful in our case. 2) More than one LSTM layers, which has been tested and is also not helpful.

\section{Conclusion and Future Works}
\label{stn:summary}
In this research, we study the server power consumption series classification problem used as a non-intrusive method for data centre energy monitoring. First we propose a new time series distance measurement termed as local time warping (LTW) and build a hybrid algorithm of the 1-nearest neighbour with LTW and the LSTM neural network. The proposed LTW distance measurement is designed to be a light weight time series measurement with local warping operations within a predefined warping index set, and it is designed to be non-commutative. LTW can be taken as the simplified version of DTW with only the warping operation (a series of ``min" operations). The LTW is proved to be better than DTW on our data set and its non-commutative feature is proved to be beneficial. Also a linear version of LTW can perform almost as good as the DTW on our data set. The proposed LTW shows that for a certain time series classification problem, it is possible to use some light weight time series distance measurement to achieve quite good classification accuracy.

Second we apply the state-of-art sequential data modelling neural network LSTM to classify the power series. Our study show that LSTM can perform well compared to 1NN-LTW with similar accuracy; however, these two algorithms have their unique different natures and the accurately classified samples of these two algorithms have significant difference. In this sense, we propose a hybrid algorithm of the two classifiers termed as LSTM/LTW, which further improves the accuracy. The proposed hybrid algorithm can achieve classification accuracy as high as 93\% in our experiments.

For the future work, one interesting problem is to study the case that the power series generated by multiple programs thus with multiple labels. The problem is especially interesting when we have the test data being the combination of different programs (such as a pair of programs (A, B)) where this special pair may not be seen in the training data, for example, the training data may only contain samples generated by program pairs like (B,C) and (A,C). In this case, the classifier should be able to recognize the new pair (A, B). Also, one can try more complicated ensemble algorithms with LSTM and other existed time series classification algorithms.





%

\bibliographystyle{IEEEtran}
\bibliography{IEEEabrv,pclassify}

\newpage

\end{document}